\documentclass[runningheads,a4paper]{llncs}

\usepackage{amssymb}
\usepackage{amsmath}
\usepackage{algorithm}
\usepackage{multirow}
\usepackage{algorithmic}
\usepackage{graphicx}

\usepackage{url}
\urldef{\mailsa}\path|{alfred.hofmann, ursula.barth, ingrid.haas, frank.holzwarth,|
\urldef{\mailsb}\path|anna.kramer, leonie.kunz, christine.reiss, nicole.sator,|
\urldef{\mailsc}\path|erika.siebert-cole, peter.strasser, lncs}@springer.com|    
\newcommand{\keywords}[1]{\par\addvspace\baselineskip
\noindent\keywordname\enspace\ignorespaces#1}

\begin{document}

\mainmatter  

\title{A Novel Combined Optical Flow Approach for Comprehensive Micro-Expression Recognition}

    \titlerunning{A Novel Combined Optical Flow Approach for MER}

%
%

\author{Vu Tram Anh Khuong \and Thi Bich Phuong Man \and Luu Tu Nguyen \and Thanh Ha Le \and Thi Duyen Ngo%
\thanks{Corresponding author: Thi Duyen Ngo (email: duyennt@vnu.edu.vn)}%
\\}
\authorrunning{Tram Anh et al.}


\institute{Faculty of Information Technology,\\
VNU University of Engineering and Technology\\}

%
%
    \toctitle{A Novel Optical Flow Approach for MER}
    \tocauthor{Tram Anh et al.}
\maketitle

\begin{abstract}  
Facial micro-expressions are brief, involuntary facial movements that reveal hidden emotions. Most Micro-Expression Recognition (MER) methods that rely on optical flow typically focus on the onset-to-apex phase, neglecting the apex-to-offset phase, which holds key temporal dynamics. This study introduces a Combined Optical Flow (COF), integrating both phases to enhance feature representation. COF provides a more comprehensive motion analysis, improving MER performance. Experimental results on CASMEII and SAMM datasets show that COF outperforms single optical flow-based methods, demonstrating its effectiveness in capturing micro-expression dynamics.  

\end{abstract}

\section{Introduction}
\label{s: introduction}
Facial expressions, categorized as macro- and micro-expressions (MEs), are essential for emotional communication. Macro-expressions are high-intensity, long-lasting, while MEs are low-intensity and last less than 0.5 seconds, making them challenging to detect. MEs reveal genuine emotions that individuals may try to conceal, playing a vital role in fields like psychology \cite{Bhushan2015}, security \cite{yan2013fast}, and behavior analysis \cite{polikovsky2010detection}. Recent advancements in deep learning have significantly improved Micro-Expression Recognition (MER), with methods like ResNet, Inception, and VGGNet being used to extract spatial features, and temporal models such as LSTM, GRU, and 3D-CNN capturing time-based changes.

MER techniques typically rely on two types of input modalities: appearance-based and motion-based methods. Appearance-based methods, like LBP-TOP \cite{pfister2011differentiating} and HOG \cite{dalal2005histograms}, extract static features from a single frame, often the apex, but are limited by their inability to capture the temporal dynamics that define MEs. Motion-based approaches, particularly optical flow, are preferred because of their ability to capture subtle facial movements over time while minimizing identity-specific features. Optical flow has shown strong performance in MER, highlighting the temporal dynamics of MEs. However, most existing optical flow methods focus solely on the onset-to-apex phase, which overlooks the critical motion information in the apex-to-offset phase.

This study proposes a novel approach that integrates optical flow from both the onset-to-apex and apex-to-offset phases into a Combined Optical Flow (COF). By combining information from both phases, our method addresses the limitations of previous approaches and provides a more comprehensive representation of micro-expression dynamics. This holistic view improves MER by capturing both the rapid rise and subtle decline of expressions, ultimately enhancing recognition accuracy and robustness.

The rest of this paper is structured as follows: Section \ref{s: proposed_method} details our method, Section \ref{s:Experiment and results} presents experimental results, and Section \ref{s:conclusion} concludes the study.

\section{Proposed method}
\label{s: proposed_method}
\begin{figure}[h]
    \centering
    \includegraphics[width=1\linewidth]{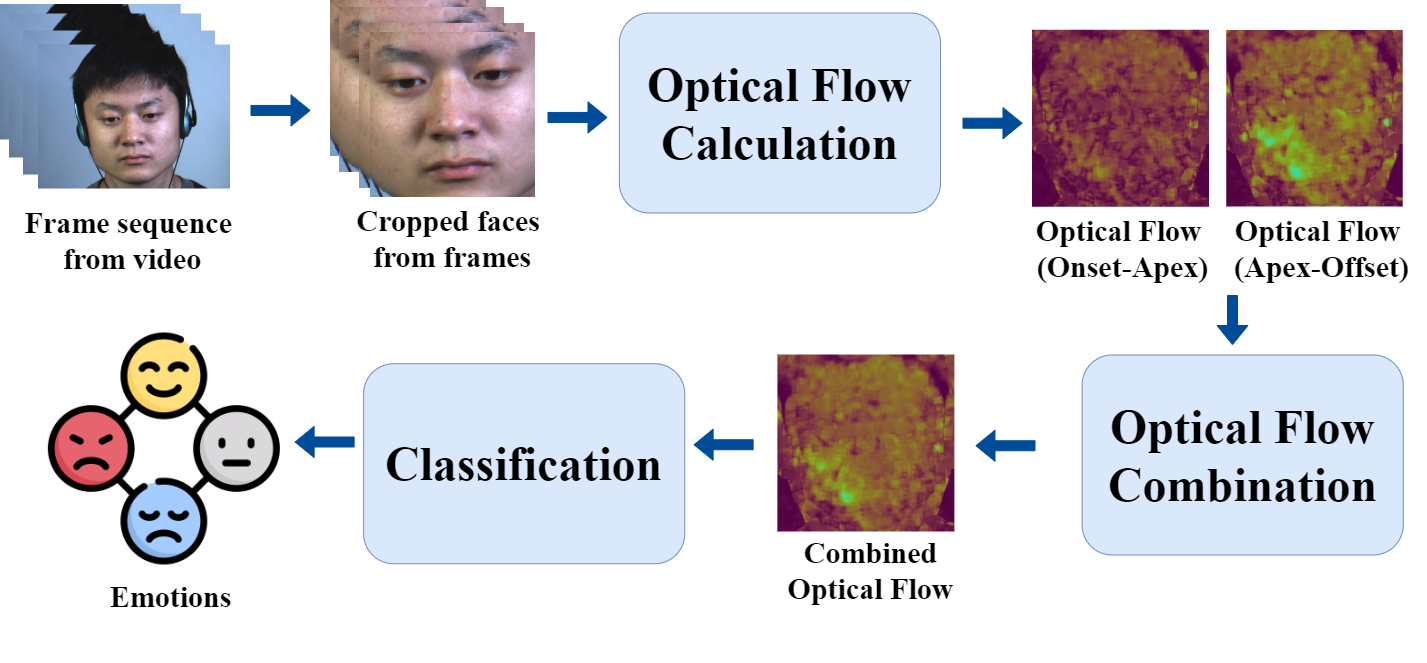}
    \caption{Pipeline for MER with Combined Optical Flow}
    \label{fig:pipline}
\end{figure}

Previous studies have primarily focused on the onset-to-apex phase, assuming a symmetrical temporal structure where the apex-to-offset phase mirrors it. However, this may oversimplify micro-expression dynamics, as apex frames in datasets are not symmetrically positioned. This paper introduces Combined Optical Flow (COF), integrating motion from both phases for a richer representation. Our pipeline (as shown in Fig. \ref{fig:pipline}) consists of: (1) extracting facial frames, (2) computing optical flow for both phases, (3) merging these flows, and (4) classifying emotions using a recognition network. Minimal preprocessing was applied to assess the model’s ability to learn directly from optical flow, avoiding added complexity or biases.

\subsection{Optical Flow Calculation}
\label{s: Optical Flow Calculation}

Optical flow measures pixel displacement between consecutive video frames to estimate object motion. This work uses the Farneback \cite{farneback} method to compute dense optical flow by analyzing changes in pixel intensities, generating a flow field that captures both magnitude and direction at each pixel. 

\subsection{Optical Flow Combination}
\label{s: Combined Optical Flow}
The combined approach for optical flow estimation is described in Algorithm \ref{alg: COF}. 
\begin{algorithm}[h]
    \caption{Optical Flow Combination}
    \label{alg: COF}
    \begin{algorithmic}[1]
        \REQUIRE \texttt{frame1} (onset), \texttt{frame2} (apex), \texttt{frame3} (offset) 
        \ENSURE Combined Optical Flow Image: \texttt{combinedOF}

        \STATE \texttt{flow1} $\gets$ \text{calcOpticalFlowFarneback}(\texttt{frame1}, \texttt{frame2})
        \STATE \texttt{flow2} $\gets$ \text{calcOpticalFlowFarneback}\texttt{(frame2, frame3)}

        \STATE \texttt{magnitude1, magnitude2} $\gets$ \text{getMagnitude}(\texttt{flow1, flow2})

        \STATE \texttt{norm\_magnitude1, norm\_magnitude2} $\gets$ \text{normalize}\texttt{(magnitude1, magnitude2)}

        \STATE \texttt{magnitude} $\gets$ \texttt{normalized\_magnitude1 + normalized\_magnitude2}

        \STATE \texttt{norm\_magnitude} $\gets$ \text{normalize}\texttt{(combined\_magnitude)}

        \STATE \texttt{combinedOF} $\gets$ \text{convertToRGB}\texttt{(norm\_magnitude)}

        \RETURN \texttt{combinedOF}
    \end{algorithmic}
\end{algorithm}

The details of the algorithm are as follows:

\begin{enumerate}
    \item \textbf{Optical Flow Calculation:}
    For each phase (onset-to-apex and apex-to-offset), dense optical flow fields are computed as described in section \ref{s: Optical Flow Calculation} to represent pixel motion magnitude and direction. 
    
   \textit{\textbf{Onset-to-Apex Phase:}} This phase captures the emergence and peak of the expression, with rapid facial contractions indicating rising intensity, essential for detecting micro-expressions.
    
    \textit{\textbf{Apex-to-Offset Phase:}} This phase reflects the decline of the micro-expression, providing insights into emotional shifts and intent, crucial for understanding the full expression cycle.
    
    \item \textbf{Magnitude Extraction:} Optical flow magnitudes are extracted from both phases to capture motion intensity changes, ensuring a more comprehensive representation of micro-expression dynamics.
    
    \item \textbf{Normalization:} Optical flow magnitudes from both phases are normalized for equal treatment, with temporal alignment preserving the correct motion sequence.
    
    \item \textbf{Optical Flow Combination:}  After normalization, magnitudes from 2 phases are summed to form a unified flow representation, creating a dense optical flow field that captures the complete motion dynamics from onset to offset.
\end{enumerate}

\subsection{Classification} 
\label{s: Model} 
The Combined Optical Flow, detailed in Section \ref{s: Combined Optical Flow}, is fed into the MER classification models. Two models are used: VGG19 \cite{Simonyan2014VeryDC}, a 19-layer deep architecture effective for feature extraction in micro-expressions, and ResNet50 \cite{he2016deep}, which employs residual learning to tackle the vanishing gradient problem and effectively capture complex motion and temporal changes.

\section{Experiment and Results}
\label{s:Experiment and results}

\subsection{Experiment}
This section compares micro-expression recognition using Optical Flow and Combined Optical Flow (COF) on CASMEII  \cite{casme} and SAMM \cite{samm} datasets with VGG19 and ResNet50. Both methods are evaluated under identical conditions to isolate the impact of input modalities, assessing the effectiveness of COF in enhancing recognition performance.

\subsubsection{Dataset}  
The Combined Optical Flow (COF) approach was tested on two benchmark datasets: CASMEII \cite{casme} (247 samples, 26 participants) and SAMM \cite{samm} (159 samples, 32 participants). After excluding labels with fewer than 10 instances, CASMEII and SAMM were reduced to five classes each. Data augmentation techniques like horizontal flipping and ±5°/±10° rotations were used. This study focuses on benchmark datasets for validation due to their consistency and widespread use in micro-expression recognition.

\subsubsection{Evaluation Metrics}
The COF method is evaluated against the optical flow baseline using a confusion matrix and accuracy, which measures the ratio of correct predictions to total samples.

\subsubsection{Implementation Details}
This study implements both the Combined Optical Flow (COF) and baseline Optical Flow methods in a structured pipeline for reproducibility. Datasets are split 80\% for training and 20\% for testing, with face regions used to compute optical flow for both phases. The baseline optical flow is computed as described in Section \ref{s: Optical Flow Calculation}, while the Combined Optical Flow method is detailed in Section \ref{s: Combined Optical Flow}. While the COF method introduces some computational overhead compared to single-phase methods, the increase is minimal and does not significantly impact training or inference times. VGG19 and ResNet50 models are fine-tuned with the Adam optimizer (learning rate $10^{-4}$, batch size 32) for 50 epochs, with all layers frozen except the last four.

\begin{figure}[h]
    \centering
    \includegraphics[width=1\linewidth]{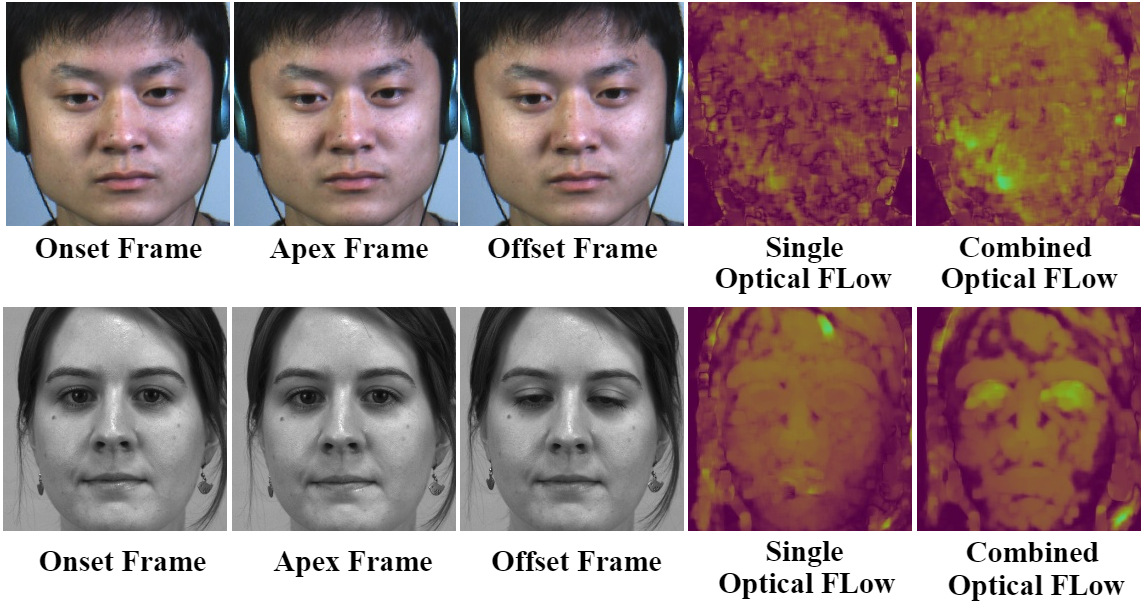}
    \caption{Visualization of Onset frame, Apex frame, Offset frame, Optical Flow (Onset-Apex), and COF from CASME-II (top row) and SAMM (bottom row) datasets. \textit{(Best viewed in color)}}
    \label{fig:OF2}
\end{figure}

\subsection{Results}
Fig. \ref{fig:OF2} compares Single Optical Flow and the proposed Combined Optical Flow (COF) alongside onset, apex, and offset frames. Visual inspection suggests that COF  captures finer-grained dynamic transitions in facial expressions, potentially enhancing MER performance over Single Optical Flow.
\begin{table}[h]
\caption{Performance Comparison of Modalities and Models on CASME II and SAMM}
\label{tab:table01}
\centering
\begin{tabular}{|l|l|c|c|}
\hline
\textbf{Input Modality} & \textbf{DCNN Model} & \textbf{CASME-II} & \textbf{SAMM} \\ \hline
                               
\multirow{2}{*}{Single Optical Flow} & VGG19    & 61.22\% & 50\% \\ \cline{2-4} 
& ResNet50 & 40.82\% & 42.86\% \\ \hline
                              
\multirow{2}{*}{\textbf{Combined Optical Flow (Ours)}} & VGG19 & \textbf{67.35\%} & \textbf{59.26\%} \\ \cline{2-4} & ResNet50   & \textbf{55.10\%} & \textbf{59.26\%} \\ \hline

\end{tabular}
\end{table}

\begin{figure}[h]
    \centering    \centerline{\includegraphics[width=1\linewidth]{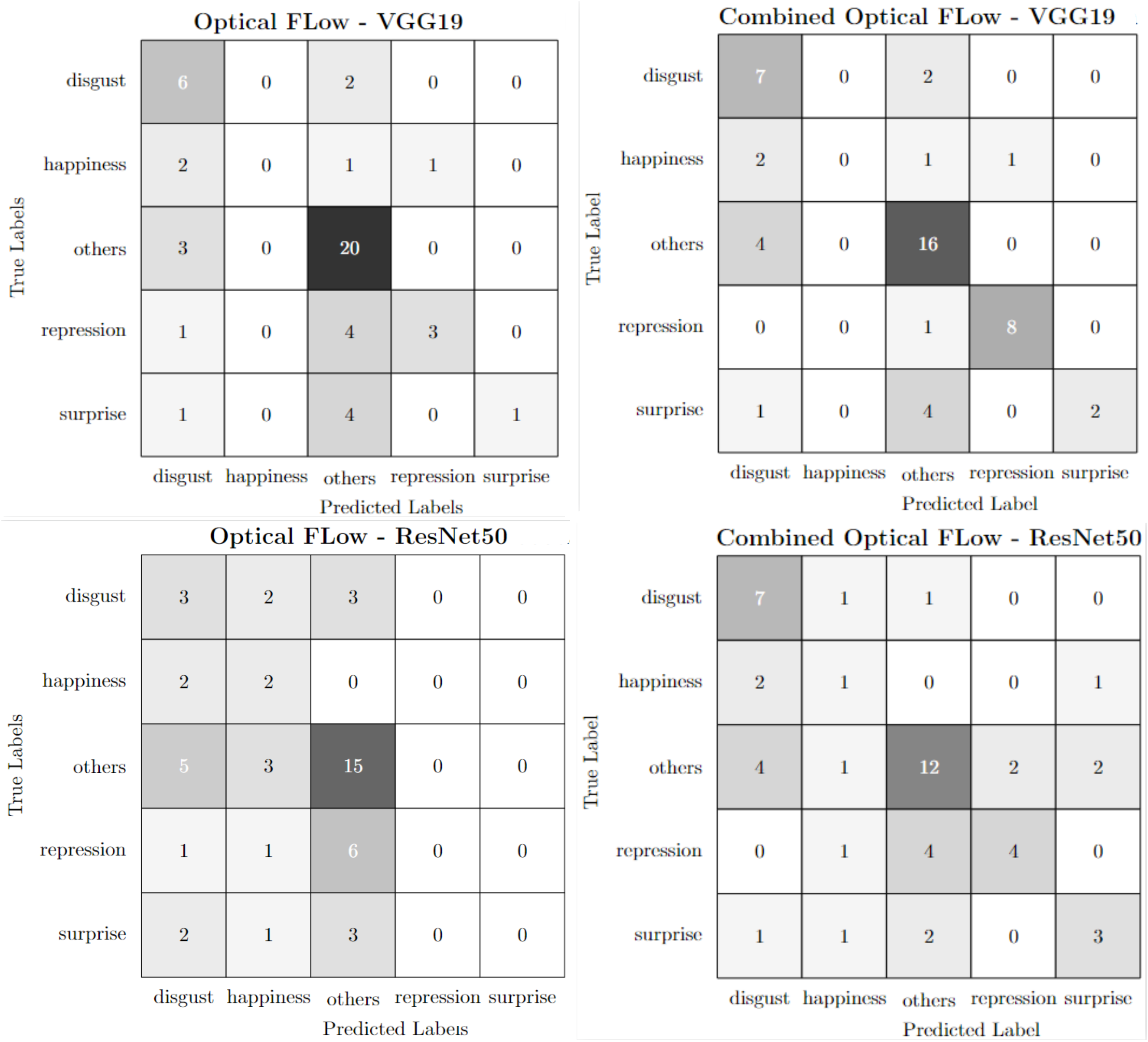}}
    \caption{Confusion Matrices for Optical Flow and Combined Optical Flow in MER}
    \label{fig:cm}
\end{figure}

The experimental results in Table \ref{tab:table01} demonstrate that the proposed Combined Optical Flow (COF) method significantly improves motion representation and classification accuracy in micro-expression recognition. COF consistently outperforms the single optical flow approach across both datasets and models, achieving 67.35\% accuracy on CASME with VGG19 and 59.26\% on SAMM with ResNet50.

Fig. \ref{fig:cm} illustrates the confusion matrices, further highlighting COF’s advantage. While the single optical flow method mainly predicts “others” and “disgust,” struggling with labels like “happiness” and “surprise,” COF achieves a more balanced distribution and better generalization. By integrating both onset-apex and apex-offset flows, COF captures a fuller motion representation, leading to higher recognition accuracy and more robust classification.

\section{Conclusion}
\label{s:conclusion}
This study introduces Combined Optical Flow (COF), a novel approach for Micro-Expression Recognition (MER) that integrates optical flow from both onset-to-apex and apex-to-offset phases. Experimental results show that COF outperforms single optical flow methods in both accuracy and representation of micro-expressions. By incorporating the often-overlooked apex-to-offset phase, this work addresses a key limitation in existing MER methods. The findings highlight COF’s potential to improve emotion recognition beyond the ``others" category, providing a strong foundation for future research and integration with advanced deep learning architectures.
\bibliographystyle{plain}
\bibliography{references}
\end{document}